\pdfoutput=1

\RequirePackage{fix-cm}
\documentclass[smallextended]{svjour3}

\smartqed

\usepackage{hyperref}
\usepackage{graphicx}
\usepackage{amsmath}
\usepackage{stmaryrd}
\usepackage{multirow}
\usepackage{float}
\usepackage{siunitx}
\usepackage{etoolbox}
\usepackage{subfig}
\usepackage{tabularx}

\renewrobustcmd{\bfseries}{\fontseries{b}\selectfont}
\renewrobustcmd{\boldmath}{}

\begin{document}

\title{Learning from a tiny dataset of manual annotations: 
a teacher/student approach for surgical phase recognition
}

\titlerunning{A teacher/student approach for surgical phase recognition}

\author{Tong Yu \and
        Didier Mutter \and
        Jacques Marescaux \and
        Nicolas Padoy
}

\institute{Tong Yu  \and Nicolas Padoy \at
          ICube, University of Strasbourg, CNRS, IHU Strasbourg, France\\
          \email{tyu@unistra.fr}
          \and
          Didier Mutter \and Jacques Marescaux \at
          University Hospital of Strasbourg, IRCAD and IHU Strasbourg, France
}

\date{}

\maketitle

\begin{abstract} \mbox{}\\
\paragraph{Purpose}Vision algorithms capable of interpreting scenes from a real-time video stream are necessary
for computer-assisted surgery systems to achieve context-aware behavior. In laparoscopic procedures
one particular algorithm needed for such systems is the identification of surgical phases, 
for which the current state of the art is a model based on a CNN-LSTM. 
A number of previous works using models of this kind have trained them in a fully supervised manner, requiring a 
fully annotated dataset. 
Instead, our work confronts the problem of 
learning surgical phase recognition in scenarios presenting 
scarce amounts of annotated data (under 25\% of all available video recordings). 
\paragraph{Methods}
We propose a teacher/student type of approach, where a strong predictor called the teacher, trained beforehand 
on a small dataset of ground truth-annotated videos, generates synthetic annotations for a larger dataset, which 
another model - the student - learns from. In our case, the teacher features a novel CNN-biLSTM-CRF architecture, 
designed for offline inference only. The student, on the other hand, is a CNN-LSTM capable of making real-time 
predictions. 
\paragraph{Results} Results for various amounts of manually annotated videos demonstrate the superiority of 
the new CNN-biLSTM-CRF predictor as well as improved performance from the CNN-LSTM trained using  
synthetic labels generated for unannotated videos.  
\paragraph{Conclusion}
For both offline and online surgical phase recognition with very few annotated recordings available, 
this new teacher/student strategy provides a valuable performance improvement by efficiently leveraging 
the unannotated data.

\end{abstract}

\section{Introduction}
\label{intro}
\paragraph{Problem statement} Laparoscopic surgery is, by nature, an abundant source of video data, which deep learning algorithms 
can leverage for surgical workflow analysis. 
One particular application that has been examined is phase recognition in surgical procedures. This task is 
essential for enabling context-aware behavior in Computer-Assisted Surgery: a system capable of identifying 
surgical phases in real time could use this information to deliver adequate notifications or warnings to a 
surgeon in the middle of an intervention. Real-time operation is highly desirable in this context: offline 
predictors may be employed as post-processing tools dedicated to recordings, for educational or archival 
purposes; but online predictors can directly tap into a live video stream of the procedure and be incorporated 
into a computer-assisted surgery system.
Producing annotations to train such predictors, however, is a 
potentially tedious process that often requires the presence of clinical experts. 
This issue creates a strong incentive for exploring semi-supervised methods, specifically tailored for 
situations in which manual annotations are only provided for a fraction of all available data.

\paragraph{Contribution}For the work presented in this paper, we direct our attention towards surgical phase 
recognition in scenarios 
of extreme manual annotation scarcity: 20 manually annotated recordings or fewer, which is 25\% or 
less of the training set's initial size in our dataset of cholecystectomy recordings. 
In order to perform phase recognition under these conditions, we propose a teacher/student approach, 
where the teacher - a model trained on a small set of manually annotated videos - generates synthetic 
annotations for a large number of videos, which, in turn, serve as training material for another model called the student.
In the context of surgical phase recognition, this is, to the best of our knowledge, the first work in which 
synthetic labels are employed. 
The teacher model presents a novel architecture that combines a Convolutional Neural Network (CNN), 
a bidirectional Long-Short-Term-Memory Network (biLSTM) and a Linear-Chain Conditional Random Field (CRF); the 
student model is a CNN-LSTM. The two models are highly complementary, with on the one 
hand a teacher exhibiting stronger predictive power but no real-time inference capabilities, and on the other 
hand a weaker model capable of performing in real time. 

Experiments are performed on \textit{cholec120}, a dataset of 120 cholecystectomy videos \cite{rsdnet}. 
Each video frame belongs to one of the 7 following phases: 
(P1) preparation, 
(P2) calot triangle dissection, 
(P3) clipping and cutting, 
(P4) gallbladder dissection, 
(P5) gallbladder packaging, 
(P6) cleaning and coagulation, 
(P7) gallbladder extraction. 
When only few ground truth annotated videos are available (e.g. 20), results show improvement for both the 
CNN-biLSTM-CRF model (75.8\% F1 score) trained only on ground truth annotated videos and for the CNN-LSTM 
trained with synthetic labels (73.2\% F1 score), as compared to the CNN-LSTM trained on ground truth annotated 
videos alone (70.2\% F1 score); this brings the performance closer to the CNN-LSTM trained on all 80 manually 
annotated videos (78.2\% F1 score).

\paragraph{Related work} Surgical phase recognition from video data has recently received much 
attention from the Computer-Assisted Interventions community, and the state-of-the-art approaches rely on deep 
learning. 
An example of early work relying on deep learning is Twinanda et. al \cite{endonet},
which features a combination of CNN and HMM. 
A staple approach for surgical phase recognition in the fully supervised 
case combines a CNN, used as a visual 
feature extractor, with a unidirectional LSTM to model temporal dependencies in the video. 
This approach was presented in \cite{andrum2cai} and the related PhD thesis \cite{andruphd}.
The author also experimented with a biLSTM, which showed increased 
performance but only functioned as an offline predictor. Jin et al. used the same type of model in \cite{jin}, 
but aggregated features extracted by the CNN from three timesteps close to each other 
before feeding the LSTM. The same author later introduced an extension in \cite{svrcnet}, using 
Prior Knowledge Inference to post-process the predictions.
In our work, we rely on a CNN-LSTM architecture for the student model as well, and a 
CNN-biLSTM for a large part of the teacher model.

Conditional Random Fields (CRF) have been featured in several works on surgical phase recognition: 
\cite{quellec_crf}
and \cite{charriere_crf}
both used them for cataract surgery videos. Lea et al. relied on CRFs for analyzing surgical training tasks in 
\cite{lea_crf_1} and \cite{lea_crf_2}. 
None of these approaches, however, have combined them with CNN-extracted visual features or deep recurrent 
temporal models.
The combination of bidirectional LSTM and CRF has been explored by \cite{lample_ner} and \cite{huang_ner} 
in the context of Named Entity Recognition - a problem involving no video data - where the objective is 
to identify the type of each word in a sentence - for instance object or location. We make use of a 
biLSTM-CRF combination in our work, in order to obtain well-structured sequences of predictions from the 
teacher. 

\cite{data_distillation} proposes a teacher/student approach on different target tasks
, namely human pose estimation and object detection. These tasks exploit single frames, 
and do not involve any temporal data. 
The method relies on synthetic labels generated by ensembling predictions over geometric transforms 
of the same input image. 

A few recent papers have approached surgical phase recognition from the semi-supervised angle, using the same 
type of CNN-LSTM architecture as in \cite{andruphd}, but with different training strategies that exploit 
unannotated videos. The core idea is to pretrain the CNN on a \textit{proxy task} for which no manual labels are 
required, in order to encourage it to learn temporally relevant features. \cite{bodenstedt} used a CNN pretrained 
on frame sorting. 
\cite{funke} trained 
the CNN to make predictions related to the temporal distance between multiple frames of a video. Experiments 
were performed on the \textit{cholec80} dataset with a full training set size of 60 and 
mini-training sets of 40 or fewer annotated videos.
Similarly, \cite{endon2n} used the prediction of Remaining Surgery Duration (RSD) as a proxy task, 
along with end-to-end training of the CNN-LSTM final model. This scheme can be challenging to implement 
due to the interplay between weight regularization and the range of the model's output, in addition 
to the taxing computational requirements of end-to-end training.
The corresponding experiments used the \textit{cholec120} dataset with a full training set size 
of 80 and mini-training sets of 64 or fewer annotated videos.


\section{Methods}
\label{sec:1}
\subsection{Overview}
\label{sec:2}

\begin{figure}[H]
  \centering
  \includegraphics[scale=0.65]{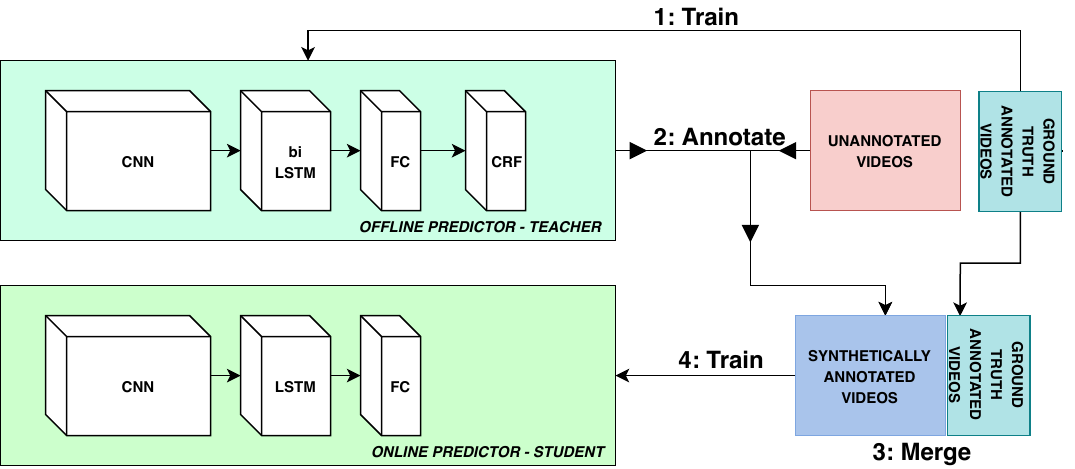}
\caption{Teacher-student approach overview}
\label{fig:1}
\end{figure}

The approach presented in this paper relies on a teacher/student dynamic between the two temporal models depicted 
in Fig.\ref{fig:1}. The teacher model is trained on a small set of manually annotated videos and generates 
synthetic labels for every video for which ground truth annotations are unavailable. The pool of synthetically 
annotated videos and manually annotated videos is then employed to train another model, namely the student.
The teacher, in our case, uses a new combination of CNN, biLSTM and CRF to achieve offline inference results 
superior to what a conventional CNN-LSTM would be capable of, 
thanks to its ability to take into account information from future timesteps (biLSTM), as well as knowledge
on transitions between phases in surgical videos (CRF). The student is a traditional CNN-LSTM, which comes with 
the major benefit of online inference. 

\subsection{Teacher model}
\label{sec:3}
We describe below each component of the CNN-biLSTM-CRF teacher model.

\subsubsection{CNN}
\label{sec:4}
The chosen CNN architecture is Resnet-50 V2, as introduced by He et al.\cite{resnet}. It serves as a 
visual feature extractor, mapping the $ 256 \times 256 \times 3 $ input RGB images to vector representations of 
size $ N_{f} = 2048 $. These feature vectors will then serve as input for the temporal model described next. 

\subsubsection{Bidirectional LSTM}
\label{sec:5}
Formally, a unidirectional LSTM maps a sequence of input vectors $(i_{0}, ..., i_{T})$ to a 
sequence of output vectors $(o_{0},..., o_{T})$ obeying
\begin{equation}
	\forall t \in \llbracket 1, T \rrbracket, (o_{t}, c_{t}, h_{t}) = q_{f}(i_{t}, c_{t-1}, h_{t-1}),
\end{equation}
where $(c_{t}), (h_{t})$ are sequences of states, with $c_{0}, h_{0}$ randomly initialized in our case. The function 
$q_{f}$ as detailed in \cite{hochreiter_lstm} is the LSTM cell; upon 
training, $q_{f}$ learns to model temporal dependencies from past timesteps to the current one.
\cite{graves_bilstm} suggests incorporating another LSTM cell, in order to model for temporal dependencies from 
future samples to the current one. This means introducing another sequence of outputs 
$(\omega_{0},..., \omega_{T})$ and an LSTM cell $ q_{b} $ such that
\begin{equation}
  \forall t \in \llbracket 0, T-1 \rrbracket, (\omega_{t}, \gamma_{t}, \eta_{t}) = 
  q_{b}(i_{t}, \gamma_{t+1}, \eta_{t+1}),
\end{equation}
again with $ (\gamma_{t}), (\eta_{t}) $ as sequences of states, with initial values $ (\gamma_{T}), (\eta_{T}) $.
The concatenated pair $(o_{t}, \omega_{t})$ therefore 
contains temporal information from both past and future events. This type of model, while very potent for offline 
inference, is by nature unsuitable for real-time operation.

\subsubsection{Linear-Chain Conditional Random Field (CRF)}
\label{sec:6}
A single fully connected layer turns the outputs of the biLSTM into logits of size $N_{c} = 7$, one for each 
surgical phase in a cholecystectomy. 
In general classification problems, each vector of logits is processed independently from the others, 
by passing it through the softmax function and computing its cross-entropy against a one-hot encoded label 
vector during training, and by taking the argmax during inference.

A better approach for our application, where consecutive predictions should be consistent with one another, is 
to process the entire sequence of logits corresponding to a given video together using a linear-chain conditional 
random field model (CRF). This can be interpreted as a form of smoothing, done in a more 
principled manner than in ad-hoc postprocessing methods such as the one featured in \cite{jin}. 

Given a sequence of logits $S = (s_{0},..., s_{T})$, for any given timestep $t$ and class index $k$ we note the
$k^{th}$ entry of $s_{t}$ as $s_{t}[k]$. Let $\Theta$ be an $N_{c} \times N_{c}$ real-valued matrix, with entries 
noted as $\Theta[i,j]$ for class indices $i, j$.
The score of any sequence of tags (i.e. predicted classes) $Y = (y_{0}, ..., y_{T})$ can then be defined as:
\begin{equation}
  C(S, Y, \Theta) = \sum_{t=0}^{T} s_{t}[y_{t}] + \sum_{t=0}^{T-1} \Theta[y_{t}, y_{t+1}] .
\end{equation}
The trainable parameter of the model is $\Theta$, called the transition matrix.
Using the definition of the score $C$, we are able to define the likelihood of a tag sequence as:
\begin{equation}
  p(Y|S) = \frac{e^{C(S, Y, \Theta)}}{\sum\limits_{U \in \llbracket 1, N_{c}\rrbracket^{T+1}}e^{C(S, U, \Theta)}},
\end{equation}
which is the entry corresponding to $Y$ in the softmax over all possible tag sequences $U$.  
We then employ $L = -log(p(Y_{true}|S))$ as our training loss, $Y_{true}$ being the ground truth tag sequence. 
During inference, the highest scoring tag sequence $Y_{opt}$ is obtained by Viterbi decoding. 

\subsection{Student model}
\label{sec:7}
The student model follows the architecture first established in \cite{endonet}. It relies on the same Resnet-50 
CNN as the teacher model for feature extraction, followed by a unidirectional LSTM.
This model is adequate for real-time predictions, since at any given timestep the output only depends on the 
current input, the previous state and the previous output of the LSTM.

\subsection{Training}
\label{sec:8}
The dataset we use for all of our experiments, \textit{cholec120}, contains 120 recordings of 
laparoscopic cholecystectomy, with an average duration of 38 minutes \cite{rsdnet}. We reserve 30 for testing and 10 
for validation, leaving $N = 80$ videos to choose from for training. In the following lines we will refer to:
\begin{equation}
  E = \left\{ (V_{0}, \mathcal{A}(V_{0})), (V_{1}, \mathcal{A}(V_{1})), ..., (V_{N-1}, \mathcal{A}(V_{N-1})) \right\}, 
\end{equation}
as the 80 manually annotated videos to choose from, $ V_{k} $ and $ \mathcal{A}(V_{k}) $ being a video and its 
ground truth annotations respectively.

Considering the voluntarily low number of annotated videos selected for training (from 20 down to 1), we sample 
3 non-overlapping mini-training sets of every 
size in order to prevent biased results coming from the selection process, and ran our
series of experiments independently for each mini-training set. In an effort to match the original training set 
in terms of featured video lengths, we divide the 80 videos into duration quartiles $Q1$ to $Q4$, then 
randomly sample videos from $Q1$, $Q2 \bigcup Q3$, and $Q4$ with a 20/60/20 ratio respectively. For 
mini-training sets of 
only one video the single choice is limited to $Q2 \bigcup Q3$ in order to avoid outliers. The mini-training 
sets are referred to as:
\begin{equation}
  E_{i, j} = \left\{ (V_{k}, \mathcal{A}(V_{k})), ... \right\},
\end{equation}
where $i \in \{1, 3, 5, 10, 20\}$ indicates the size of the set of ground truth labeled videos employed, and 
$j \in \{0, 1, 2\}$ is the index for the first, second or third experiment for that particular size. 

In all of the following experiments, the Resnet-50 CNN is initialized with ImageNet pretrained 
weights\footnote{sourced from \url{https://github.com/tensorflow/models/tree/master/research/slim}}.
Test accuracy on ImageNet for those weights reaches 75.6\%. The teacher model's CNN is first pretrained with 
only one fully connected layer on top, on $ E_{i, j} $, directly 
for phase recognition. Weights from the first and second blocks of Resnet are frozen. 
Data augmentation is applied using $ \pm 16^{\circ}$ rotations and 4 different translations.
Visual feature vectors are then extracted from $ E_{i, j} $ using the pretrained CNN.

The biLSTM - CRF is then trained end-to-end on the extracted features with untruncated backpropagation through 
time across the entire video, with the loss function defined in section \ref{sec:6} similarly to 
\cite{lample_ner} and \cite{huang_ner}. 

Using the trained biLSTM - CRF, we generate new annotations for videos in $ E \backslash E_{i, j} $. This 
leads to the set of videos with synthetic annotations: 
\begin{equation}
	F_{i,j} = \left\{ (V_{k}, \widehat{\mathcal{A}_{i,j}}(V_{k})), ... \right\}.
\end{equation}
We then define $G_{i,j} = E_{i, j} \bigcup F_{i, j}$, which contains all videos from the original training set, 
and combines a small set of manual annotations with a majority of synthetic labels.
The student CNN-LSTM model is trained in the same two-step manner, this time on $G_{i, j}$.

Hyperparameters used for training are detailed in table \ref{tab:1}.
All experiments are performed using Tensorflow with the Adam optimizer\cite{adam}, 
on servers fitted with Nvidia 1080Ti GPUs.

\begin{table}[H]
	\caption{Table detailing hyperparameters for the featured experiments}
	\label{tab:1}
	\setlength{\tabcolsep}{1.5pt}
	\begin{minipage}[t]{.5\linewidth}
		\centering
		\begin{tabular}[t]{|p{4cm}|p{1cm}|} 
			\hline
        \multicolumn{2}{|c|}{\textbf{CNN pretraining}} \\
        \hline
			  Learning rate  & $5 \cdot 10^{-5}$ \\ 
			  \# of epochs   & 27                \\ 
			  Minibatch size & 32                \\
			  Weight decay   & $5 \cdot 10^{-4}$ \\ 
			  \hline
        \multicolumn{2}{|c|}{\textbf{LSTM}} \\
        \hline

        Learning rate              & $5 \cdot 10^{-5}$ \\ 
        \# of epochs               & 350               \\ 
        State size                 & 128               \\ 
        Dropout                    & 0.3               \\ 
        Weight decay               & $5 \cdot 10^{-4}$ \\ 
			  &                   \\ 
			\hline\noalign{\smallskip}
		\end{tabular}
	\end{minipage}
	\begin{minipage}[t]{.5\linewidth}
		\centering
    \begin{tabular}[t]{|p{4cm}|p{1cm}|} 
      \hline
      \multicolumn{2}{|c|}{\textbf{CRF}} \\
        \hline
        Learning rate  & $5 \cdot 10^{-5}$ \\ 
        \# of epochs   & 350               \\ 
        Weight decay   & $5 \cdot 10^{-3}$ \\ 
        &                   \\ 

        \hline
        \multicolumn{2}{|c|}{\textbf{BiLSTM/BiLSTM-CRF}} \\
			\hline
			  Learning rate (biLSTM)     & $1 \cdot 10^{-3}$ \\ 
			  Learning rate (biLSTM-CRF) & $1 \cdot 10^{-4}$ \\ 
			  \# of epochs               & 350               \\ 
			  State size                 & $2 \cdot 64$      \\ 
			  Dropout                    & 0.4               \\ 
			  Weight decay               & $5 \cdot 10^{-4}$ \\ 
			\hline\noalign{\smallskip}
		\end{tabular}
	\end{minipage}
\end{table}

\subsection{Complementary studies}
\label{sec:9}

\subsubsection{Ablation studies}
\label{sec:10}
In order to demonstrate the need for every component in the teacher model, we conduct a series of ablation 
studies by training and evaluating the following models: \textbf{(M1)} CNN (obtained from the pretraining step), 
\textbf{(M2)} CNN-CRF, \textbf{(M3)} CNN-unidirectional LSTM, \textbf{(M4)} CNN-biLSTM


Temporal models M2 to M4 are trained in the same two-step manner as the proposed CNN-biLSTM-CRF model, referred 
to as M5.

\subsubsection{Full supervision}
\label{sec:11}
In order to provide comparison points with the fully supervised approach, the teacher model along with every 
model featured in the ablation studies are also trained on the original set of 80 manually 
annotated videos.

\subsubsection{Self-learning of the teacher model}
\label{sec:12}
The only student model mentioned so far is the CNN-unidirectional LSTM, due to its real-time inference 
capabilities. Another interesting possibility is to also use a CNN-biLSTM-CRF as the student, 
in order to obtain a stronger offline predictor.


\section{Results \& discussion}
\label{sec:13}
To demonstrate the benefits of our approach, we have conducted a total number of 125 experiments, 
counting all mini-training set sizes and all the models featured in the complementary studies. To test our 
models, we reported their accuracy, precision, recall and F1 score during inference 
on the test set. Unless otherwise specified, precision, recall and F1 are averaged over the 7 classes. 
For every metric at a given mini-training set size, we provide the mean and standard 
deviation over the 3 repeats of the corresponding experiment.

\subsection{Teacher model performance}
\label{sec:14}
Results for every mini-training set size, the proposed teacher model (M5) and models from the ablation study 
(M1 to M4) are shown in table \ref{tab:3}. Directly applying the CRF after the CNN (M2) 
yields poor results, likely due to temporal noise affecting the logits emitted by the CNN. 
Temporal models trained on a single video are severely affected by overfitting, and therefore 
also exhibit subpar performance. With 3 or more manually annotated videos, however, 
the biLSTM and the CRF deliver significant performance improvements. 

\begin{table}[t]
  \caption{Ablation study for the teacher model, accuracy and average F1}
  \centering
	\label{tab:3}
	\small
	\begin{tabular}{ccccccc|c}
		\hline
		  &     & 1               & 3                & 5                & 10               & 20              & 80\\
		\hline
		\multirow{2}{2em}{M1} 
		  & Acc & $40.8 \pm 2.6 $ & $ 57.3 \pm 1.5 $ & $ 62.2 \pm 1.2 $ & $ 67.4 \pm 1.5 $ & $ 71 \pm 1.9$&  $75.3$ \\
		  & F1  & $23 \pm 4.3 $   & $ 39.9 \pm 4.1 $ & $ 48.8 \pm 2.6 $ & $ 56 \pm 1.9 $   & $ 59.1 \pm 1.6$& $63.8$\\
		\hline
		\multirow{2}{2em}{M2}
		  & Acc & $41 \pm 3.5 $   & $ 58 \pm 1.2 $   & $ 63 \pm 0.6 $   & $ 65 \pm 0.3 $   & $ 72.5 \pm 1.9$& $80.5$\\
		  & F1  & $22.1 \pm 5.2 $ & $ 40.7 \pm 4.6 $ & $ 48.5 \pm 3.3 $ & $ 54.3 \pm 1.3 $ & $ 59.9 \pm 1.8$& $69.7$\\
		\hline
		\multirow{2}{2em}{M3}
		  & Acc & $40.1 \pm 4.7 $ & $ 65.4 \pm 5.7 $ & $ 72.3 \pm 0.9 $ & $ 74.7 \pm 2.6 $ & $ 80.5 \pm 0.7$& $86.3$\\
		  & F1  & $5 \pm 2.9 $    & $ 47.4 \pm 9.9 $ & $ 57.9 \pm 1.8 $ & $ 62.7 \pm 3 $   & $ 70.2 \pm 1.3$& $78.2$\\
		\hline
		\multirow{2}{2em}{M4}
		  & Acc & $36.5 \pm 4.7 $ & $ 66.3 \pm 3.7 $ & $ 73.7 \pm 1.6 $ & $ 76.6 \pm 3 $   & $ 82.6 \pm 0.6$& $88.4$\\
		  & F1  & $2.9 \pm 3 $    & $ 47.9 \pm 8.5 $ & $ 60.1 \pm 5 $   & $ 67 \pm 3.8 $   & $ 74.5 \pm 1.3$& $81.7$\\
		\hline
		\multirow{2}{2em}{\textbf{M5}}
		  & Acc & $40.1 \pm 3.1 $ & $ 71.1 \pm 4.6 $ & $ 76.2 \pm 0.3 $ & $ 78.5 \pm 2.6 $ & $ 84.1 \pm 1$ & $89.5$\\
		  & F1  & $21.1 \pm 11 $  & $ 55.3 \pm 9 $   & $ 65.8 \pm 1.2 $ & $ 69.7 \pm 1.3 $ & $ 75.8 \pm 1.5$ & $82.5$\\
		\hline
	\end{tabular}
\end{table}

\begin{figure}[H]
  \caption{Ablation study models performance as a function of the number of manually annotated videos employed}
  \label{fig:2}
	\centering
	\subfloat[Accuracy]{
    \includegraphics[scale=0.28]{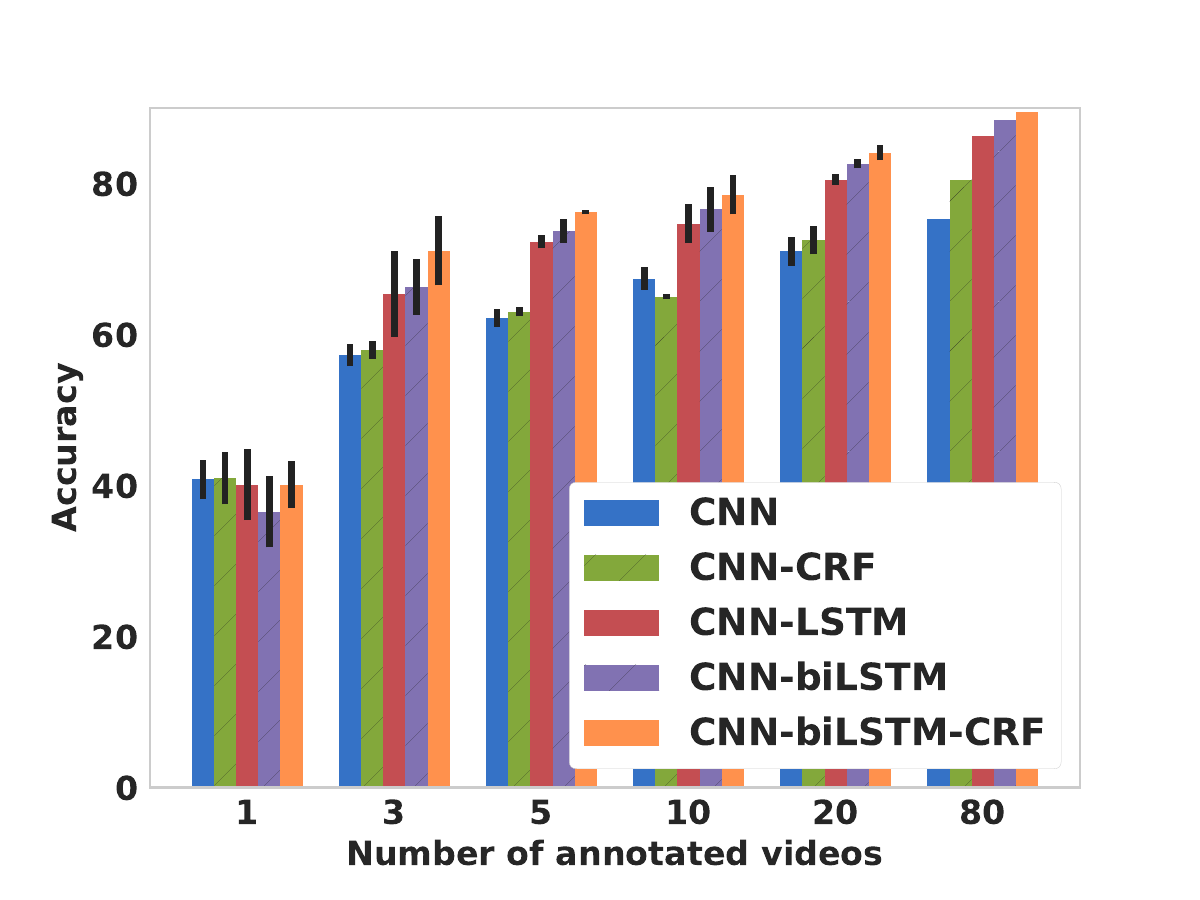}
  }
  \hspace{0.01em}
  \subfloat[F1 score]{
	  \includegraphics[scale=0.28]{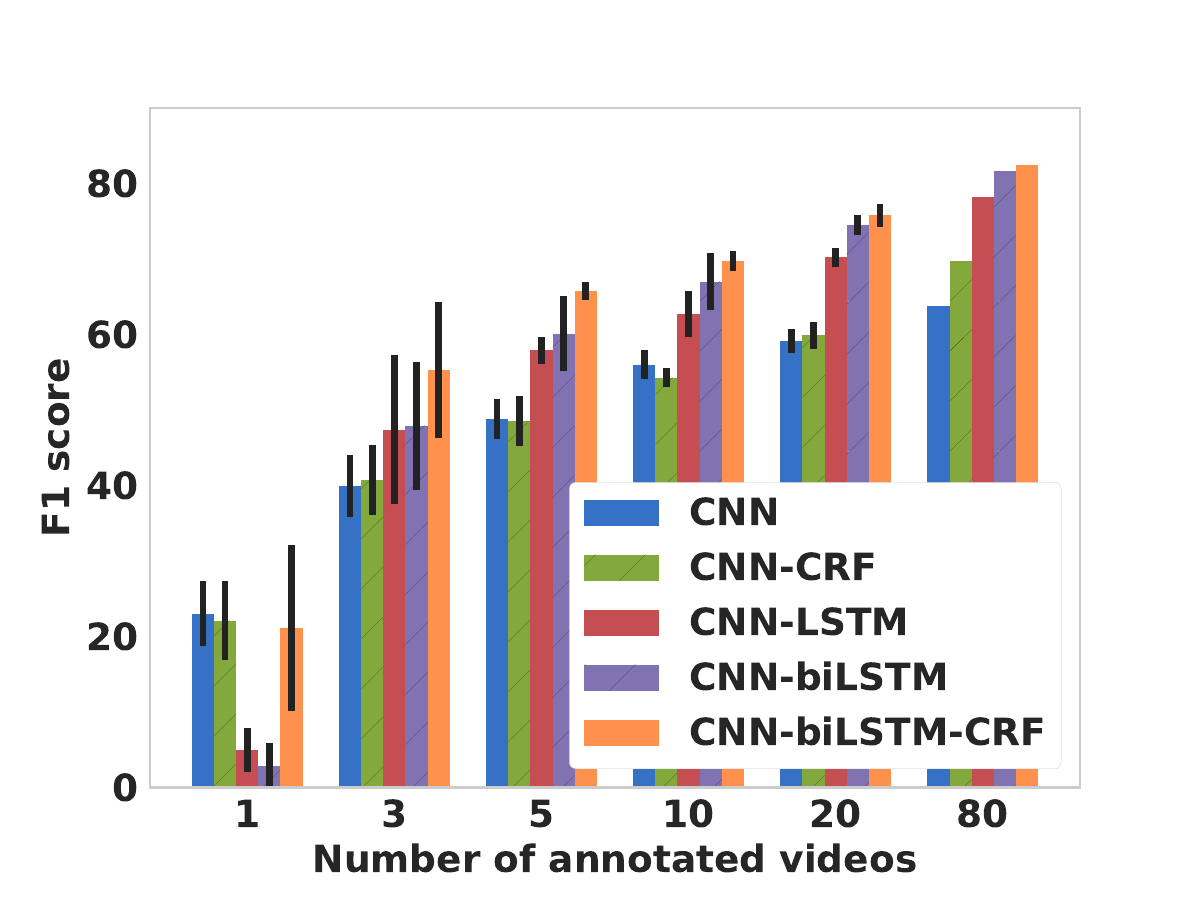}
  }
\end{figure}

The CNN-biLSTM (M4) achieves stronger performance than the CNN-unidirectional LSTM (M2), although not as 
much as the full CNN-biLSTM-CRF model (M5), which is consistently the best performer (Fig.
\ref{fig:2}). This is observed 
on all mini-training set sizes except for single videos. As expected, increasing the number of videos improves 
all global metrics - accuracy, average F1, average precision, average recall - although per-phase precision and 
recall may fluctuate (table \ref{tab:4}). This establishes the CNN-biLSTM-CRF model as the strongest predictor, 
and therefore the best suited for the role of teacher.

\begin{figure}[H]
  \centering
  \caption{Top 3 best and top 3 worst examples of predictions by time on full videos, from models trained on 20 videos}
  \label{fig:3}
  \includegraphics[scale=0.15]{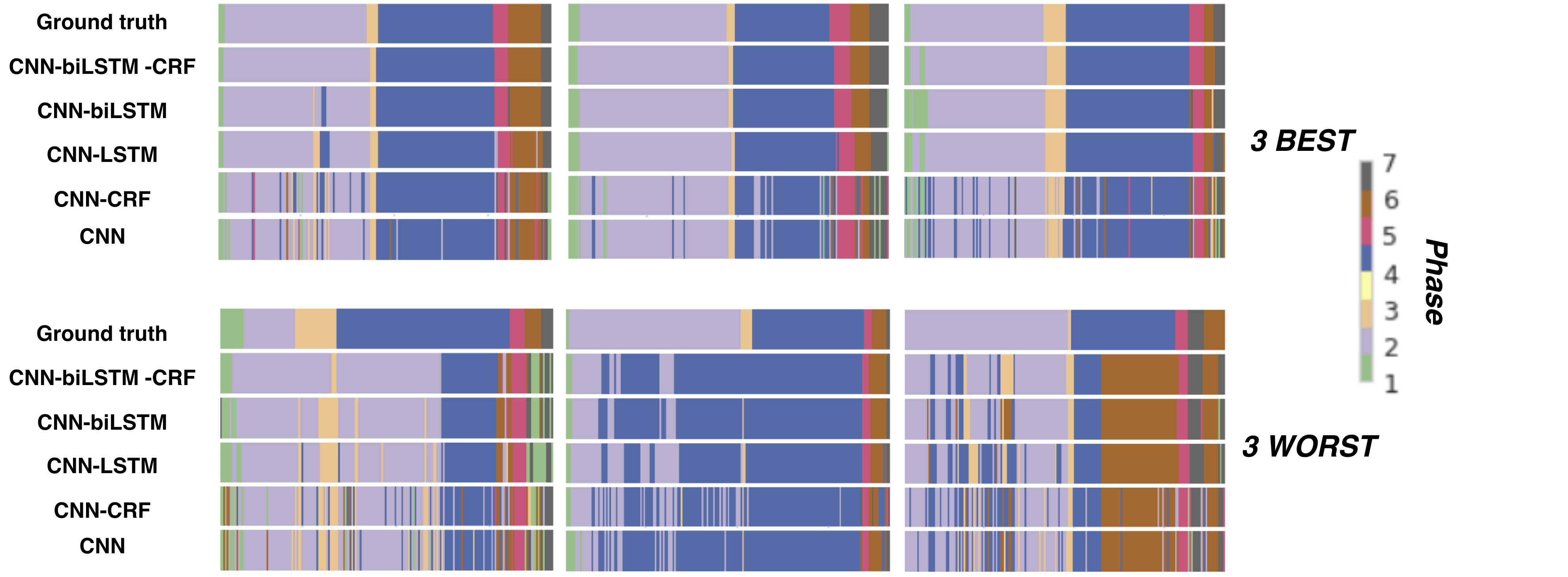}
\end{figure}

\begin{table}[t]
  \caption{Per-phase precision and recall, CNN-biLSTM-CRF model}
  \centering
	\label{tab:4}
	\small
	\begin{tabular}{ccccccc|c}
		\hline
		  &     & 1                & 3                 & 5                 & 10                & 20               & 80\\
		\hline
		\multirow{2}{2em}{P1}
		  & Pre & $5.5 \pm 8.4 $   & $ 59.6 \pm 0.8 $  & $ 52 \pm 14.3 $   & $ 56.9 \pm 14.7 $ & $ 68.4 \pm 8.3	$& $86.6$\\
		  & Rec & $56.6 \pm 26.7 $ & $ 77.4 \pm 14.8 $ & $ 51.7 \pm 4.8 $  & $ 54.4 \pm 14.1 $ & $ 92.6 \pm 4.5$ & $96.4$\\
		\hline
		\multirow{2}{2em}{P2}
		  & Pre & $48.3 \pm 42.3 $ & $ 48.1 \pm 10.5 $ & $ 80.1 \pm 3.7 $  & $ 77.1 \pm 0.9 $  & $ 58.8 \pm 10	$ & $68.9$\\
		  & Rec & $24.8 \pm 23.7 $ & $ 83.5 \pm 10.7 $ & $ 84.7 \pm 3.1 $  & $ 90.6 \pm 5 $    & $ 87.2 \pm 2.2$ & $83.4$\\
		\hline
		\multirow{2}{2em}{P3}
		  & Pre & $36.7 \pm 25 $   & $ 76.9 \pm 12.6 $ & $ 62.6 \pm 17 $   & $ 80.4 \pm 11.5 $ & $ 93.4 \pm 3.9	$& $96.5$\\
		  & Rec & $50.5 \pm 44.9 $ & $ 64.3 \pm 15.8 $ & $ 39.4 \pm 26.1 $ & $ 31.6 \pm 10.9 $ & $ 75.9 \pm 2.9$ & $79.9$\\
		\hline
		\multirow{2}{2em}{P4}
		  & Pre & $61.3 \pm 10.4 $ & $ 75.4 \pm 3.2 $  & $ 81.1 \pm 5.2 $  & $ 86.3 \pm 6.7 $  & $ 83.1 \pm 3.3	$& $89.4$\\
		  & Rec & $45.9 \pm 45.2 $ & $ 50.1 \pm 29 $   & $ 84.4 \pm 4.8 $  & $ 80.4 \pm 1.8 $  & $ 70.3 \pm 8.8$ & $72.7$\\
		\hline
		\multirow{2}{2em}{P5}
		  & Pre & $0 \pm 0 $       & $ 10.9 \pm 15.5 $ & $ 74.4 \pm 3.1 $  & $ 77.7 \pm 0.8 $  & $ 42.6 \pm 5.1	$& $53.5$\\
		  & Rec & $12.9 \pm 11.6 $ & $ 71.8 \pm 16.2 $ & $ 83.3 \pm 1.7 $  & $ 84 \pm 3.1 $    & $ 83.2 \pm 1.4$ & $88.2$\\
		\hline
		\multirow{2}{2em}{P6}
		  & Pre & $3.7 \pm 6.4 $   & $ 38.7 \pm 29.5 $ & $ 80.4 \pm 2.1 $  & $ 76.2 \pm 4.2 $  & $ 85.1 \pm 2.8	$& $91.5$\\
		  & Rec & $11.3 \pm 16.6 $ & $ 71 \pm 11.9 $   & $ 46.7 \pm 7.6 $  & $ 61.5 \pm 5.4 $  & $ 81.2 \pm 3.6$ & $91.5$\\
		\hline
		\multirow{2}{2em}{P7}
		  & Pre & $60.3 \pm 44.7 $ & $ 82.1 \pm 8.5 $  & $ 64.1 \pm 7.9 $  & $ 70.8 \pm 1.6 $  & $ 87.1 \pm 3.6	$& $94.1$\\
		  & Rec & $50.3 \pm 38.9 $ & $ 64.6 \pm 8.3 $  & $ 77.7 \pm 10.1 $ & $ 77.8 \pm 9.5 $  & $ 80 \pm 4.1$   & $77.3$\\
		\hline
		\multirow{2}{2em}{Avg}
		  & Pre & $26.4 \pm 4.5 $  & $ 62 \pm 6.1 $    & $ 70.4 \pm 4.2 $  & $ 75.1 \pm 1.6 $  & $ 75.8 \pm 1.4	$& $82.6$\\
		  & Rec & $40.5 \pm 18.5 $ & $ 62.9 \pm 7.5 $  & $ 66.8 \pm 4.7 $  & $ 68.6 \pm 0.5 $  & $ 79.8 \pm 1.3$ & $84.5$\\
		\hline
	\end{tabular}
\end{table}
To qualitatively appreciate improvements from the new model, the predictions on six videos from the test set are 
presented in Fig.\ref{fig:3}: the top three and bottom three of the CNN-biLSTM-CRF ranked by accuracy. 
The CNN-biLSTM-CRF makes the most sensible predictions with respect to the chronological order 
between the phases; it more specifically avoids inferring incorrect phases in short isolated bursts as the biLSTM 
sometimes does.

\begin{table}
	\caption{Teacher-completed annotation set metrics}
	\centering
	\label{tab:5}
	\small
	\begin{tabular}{ccccccc}
		\hline
		             &     & 1                         & 3                        & 5                        & 10                       & 20                       \\
		\hline
		\multirow{2}{2em}{P1}
		             & Pre & $ 44.4 \pm 33.4 $ & $ 63.5 \pm 10.5 $ & $ 68.8 \pm 10.3 $ & $ 78.2 \pm 5.3 $ & $ 85.9 \pm 5.5 $ \\
		             & Rec & $ 4.6 \pm 5 $ & $ 55.2 \pm 3.4 $ & $ 47.9 \pm 2.8 $ & $ 52.4 \pm 12.7 $ & $ 73.5 \pm 4.3 $ \\
		\hline
		\multirow{2}{2em}{P2}
		             & Pre & $ 60.9 \pm 4.7 $ & $ 81.1 \pm 1.9 $ & $ 86 \pm 2.5 $ & $ 83.1 \pm 2.3 $ & $ 90.3 \pm 1.6 $ \\
		             & Rec & $ 35.4 \pm 17.2 $ & $ 77.4 \pm 5.7 $ & $ 79.1 \pm 6.6 $ & $ 88.9 \pm 0.3 $ & $ 91.6 \pm 3 $ \\
		\hline
		\multirow{2}{2em}{P3}
		             & Pre & $ 79.7 \pm 28.8 $ & $ 67.9 \pm 9.4 $ & $ 66.2 \pm 18.7 $ & $ 76.3 \pm 5.4 $ & $ 84.3 \pm 2.5 $ \\
		             & Rec & $ 0.6 \pm 0.2 $ & $ 19.5 \pm 12.3 $ & $ 47 \pm 23.2 $ & $ 40.9 \pm 7 $ & $ 60 \pm 5.8 $ \\
		\hline
		\multirow{2}{2em}{P4}
		             & Pre & $ 53.6 \pm 18.6 $ & $ 72.7 \pm 10.5 $ & $ 74.5 \pm 4 $ & $ 83.2 \pm 0.6 $ & $ 87.4 \pm 3.6 $ \\
		             & Rec & $ 60.8 \pm 36.9 $ & $ 84.6 \pm 5.2 $ & $ 89.2 \pm 2.9 $ & $ 86.4 \pm 4.4 $ & $ 92.5 \pm 1.5 $ \\
		\hline
		\multirow{2}{2em}{P5}
		             & Pre & $ 87.2 \pm 11.4 $ & $ 66.6 \pm 14.9 $ & $ 77.6 \pm 1.9 $ & $ 82.8 \pm 3.1 $ & $ 83.9 \pm 2.9 $ \\
		             & Rec & $ 23.2 \pm 16.5 $ & $ 78.6 \pm 9.1 $ & $ 81.7 \pm 1 $ & $ 81.1 \pm 4.2 $ & $ 88.3 \pm 2.1 $ \\
		\hline
		\multirow{2}{2em}{P6}
		             & Pre & $ 43.4 \pm 39.5 $ & $ 70.1 \pm 9.1 $ & $ 71.8 \pm 7.5 $ & $ 76.9 \pm 4.2 $ & $ 87.3 \pm 1.2 $ \\
		             & Rec & $ 43.9 \pm 38.9 $ & $ 52.2 \pm 24.1 $ & $ 51.2 \pm 10.1 $ & $ 71.2 \pm 10.1 $ & $ 82 \pm 4 $ \\
		\hline
		\multirow{2}{2em}{P7}
		             & Pre & $ 58.1 \pm 31.3 $ & $ 62.4 \pm 5 $ & $ 63.3 \pm 9.2 $ & $ 71.8 \pm 5.9 $ & $ 84.4 \pm 4.3 $ \\
		             & Rec & $ 13.4 \pm 14.8 $ & $ 72.3 \pm 8.1 $ & $ 77.6 \pm 7.9 $ & $ 81.2 \pm 1.8 $ & $ 84.2 \pm 2.6 $ \\
		\hline
		\multirow{2}{2em}{Avg}
		             & Pre & $ 61.3 \pm 3.8 $ & $ 69.2 \pm 2 $ & $ 72.6 \pm 5.3 $ & $ 78.9 \pm 0.8 $ & $ 86.2 \pm 1.3 $ \\
		             & Rec & $ 26 \pm 15.7 $ & $ 62.8 \pm 5.3 $ & $ 67.7 \pm 4.7 $ & $ 71.7 \pm 2.3 $ & $ 81.7 \pm 0.6 $ \\
		\hline
		\hline
		\textbf{Acc} &     & \bfseries 39.2 $\pm$ 4.7  & \bfseries  73.1 $\pm$ 3.1  & \bfseries  76.6 $\pm$ 1.2  & \bfseries 81.6 $\pm$ 1  & \bfseries 88 $\pm$ 1.2  \\
		\hline
		\textbf{F1}  &     & \bfseries 23.5 $\pm$ 3.2  & \bfseries 62 $\pm$ 4.7  & \bfseries 67.6 $\pm$ 0.8  & \bfseries 73.7 $\pm$ 2.1  & \bfseries 83.5 $\pm$ 1.1  \\
		\hline
	\end{tabular}
\end{table}

\subsection{Quality of the teacher-completed annotation sets}
\label{sec:15}

In order to confirm the quality of the material the student model learns from, we ran our metrics on 
the annotations from 
$G_{i,j}$, which mix ground truth annotations and teacher-generated annotations. 
Results are shown in Table \ref{tab:5}. 
As expected, teachers trained on more manually annotated videos produce better annotations. 
The presence of more manually annotated videos in the $G_{i,j}$ sets also contributes to 
greater overall label quality, e.g. 83.5 \% F1 score for $G_{20, j}$ on average. 
Per-phase results indicate stronger 
performance on phases P2 (90.3\% precision, 91.6 \% recall for 20 manually annotated videos) and P4 
(87.4\% precision, 92.5 \% recall), which are usually the most prevalent. 
Despite uneven results across phases, $G_{i,j}$ sets obtained from 3 or more manually annotated 
videos appear to be overall serviceable for training a student model.

\subsection{Student model performance}
\label{sec:16}
To appreciate the benefits of using our synthetic labels, we compare in the first two groups of rows of 
Table \ref{tab:6} the CNN-LSTM only 
trained with few manually annotated videos against the CNN-LSTM trained with these same videos and annotations 
plus the videos annotated by the teacher. 
Results for a single ground truth-annotated video, as one can expect from the results of the corresponding 
teacher models, are quite poor. Even though the $G_{1, j}$ sets contain 80 times as many videos as the 
$E_{1, j}$ used for the teachers, the quality of the annotations, as shown in section \ref{sec:15}, is 
extremely low. CNN-LSTM models trained on those all exhibit sub-50\% performance on every global metric, despite 
showing some improvement compared to the CNN-LSTM trained on $E_{1, j}$. Decent results are observed starting from 3 videos, with a 2.9 to 8.8 point increase in accuracy when adding 
synthetic annotations, and similar increase in F1 score.

With 80 ground truth-annotated videos, accuracy and F1 reach 86.3\% and 78.2\% respectively (Table \ref{tab:6}). 
Therefore, the use of synthetic annotations roughly reduces the performance gap between using 20 and 80 ground 
truth-annotated videos by half.

\begin{table}
	\caption{Performance with and without teacher-generated annotations.}
	\centering
	\label{tab:6}
	\small
	\begin{tabular}{ccccccc}
		\hline
		  &     & 1                & 3                & 5                & 10               & 20              \\
		\hline
		\multirow{4}{6em}{CNN-LSTM, no teacher}
		  & Acc & $40.1 \pm 4.7 $  & $ 65.4 \pm 5.7 $ & $ 72.3 \pm 0.9 $ & $ 74.7 \pm 2.6 $ & $ 80.5 \pm 0.7$ \\
		  & Pre & $20.7 \pm 1.9 $  & $ 51.7 \pm 7.1 $ & $ 60.4 \pm 2.9 $ & $ 64.1 \pm 1.5 $ & $ 71.1 \pm 0.7$ \\
		  & Rec & $17.5 \pm 7.5 $  & $ 58.6 \pm 8.2 $ & $ 64 \pm 5.5 $   & $ 68 \pm 3.1 $   & $ 64.1 \pm 2.7$ \\
		  & F1  & $5 \pm 2.9 $     & $ 47.4 \pm 9.9 $ & $ 57.9 \pm 1.8 $ & $ 62.7 \pm 3 $   & $ 70.2 \pm 1.3$ \\
		\hline
		\multirow{4}{6em}{CNN-LSTM, with teacher}
		  & Acc & $42.1 \pm 6.3 $  & $ 74.6 \pm 3.6 $ & $ 77.7 \pm 0.8 $ & $ 79.1 \pm 0.9 $ & $ 83.4 \pm 0.3$ \\
		  & Pre & $24.6 \pm 5.6 $  & $ 63 \pm 5 $     & $ 65.8 \pm 5.1 $ & $ 66.6 \pm 1.7 $ & $ 73.5 \pm 1.2$ \\
		  & Rec & $38.9 \pm 19.4 $ & $ 65.1 \pm 7.3 $ & $ 72.3 \pm 5.2 $ & $ 74.7 \pm 1.7 $ & $ 76.8 \pm 0.7$ \\
		  & F1  & $14.5 \pm 13.5 $ & $ 56.1 \pm 8.6 $ & $ 64.5 \pm 2.9 $ & $ 66.9 \pm 3.1 $ & $ 73.2 \pm 0.8$ \\
    \hline
    \hline
		\multirow{4}{6em}{CNN-biLSTM-CRF, no teacher}
		  & Acc & $40.1 \pm 3.1 $  & $ 71.1 \pm 4.6 $ & $ 76.2 \pm 0.3 $ & $ 78.5 \pm 2.6 $ & $ 84.1 \pm 1$   \\
		  & Pre & $26.4 \pm 4.5 $  & $ 62 \pm 6.1 $   & $ 70.4 \pm 4.2 $ & $ 75.1 \pm 1.6 $ & $ 75.8 \pm 1.4$ \\
		  & Rec & $40.5 \pm 18.5 $ & $ 62.9 \pm 7.5 $ & $ 66.8 \pm 4.7 $ & $ 68.6 \pm 0.5 $ & $ 79.8 \pm 1.3$ \\
		  & F1  & $21.1 \pm 11 $   & $ 55.3 \pm 9 $   & $ 65.8 \pm 1.2 $ & $ 69.7 \pm 1.3 $ & $ 75.8 \pm 1.5$ \\
		\hline
		\multirow{4}{6em}{CNN-biLSTM-CRF, with teacher}
		  & Acc & $43.1 \pm 7.5 $  & $ 74.9 \pm 4.6 $ & $ 78.7 \pm 1 $   & $ 80.8 \pm 0.8 $ & $ 86.3 \pm 1$   \\
		  & Pre & $26.3 \pm 4.1 $  & $ 64.2 \pm 5.8 $ & $ 67.4 \pm 5.9 $ & $ 70.2 \pm 2.7 $ & $ 78 \pm 1.4$   \\
		  & Rec & $38.8 \pm 16.5 $ & $ 65 \pm 7.3 $   & $ 73.8 \pm 5 $   & $ 76.9 \pm 1.9 $ & $ 81.7 \pm 0.5$ \\
		  & F1  & $18.2 \pm 10.9 $ & $ 55.7 \pm 9.7 $ & $ 66.8 \pm 2.7 $ & $ 69.9 \pm 4 $   & $ 78.1 \pm 1$   \\
		\hline
	\end{tabular}
\end{table}
\subsection{Self-learning of the teacher model}
\label{sec:17}

By using the $G_{i, j}$ sets to train a new CNN-biLSTM-CRF model, the offline performance increases even 
further (Table \ref{tab:6}, last 2 groups of rows) - except for the 1-video case, where performance actually degrades. 
Results with 20 ground truth-annotated videos are particularly notable, as they match those obtained 
with 80 ground truth-annotated videos from the CNN-LSTM model (86.3\% accuracy, 78.2\% F1).


\section{Conclusion}
The work presented in this paper shows the superior performance of the new CNN-biLSTM-CRF teacher architecture 
compared to the previous CNN-LSTM used for surgical phase recognition. Although this model is restricted to 
offline inference, we also propose a teacher/student strategy that leverages the new model for real-time 
prediction, by exploiting it as a source of synthetic annotations for a CNN-LSTM student model.

Experimental results, obtained using manual annotations for 25\% or less of all available training data, 
show serious potential for scaling surgical phase recognition to a large number of videos while alleviating 
the burden of collecting manual annotations. The performance deficit between scenarios with 25\% and 100\% 
ground truth annotation availability is halved for a CNN-LSTM online prediction model when adding synthetic 
labels from the teacher. When swapping the student for a CNN-biLSTM-CRF offline prediction model, the gap 
is fully closed. Many other types of surgical procedures than cholecystectomy, 
for which large amounts of annotated data are not yet available, might be able to benefit from 
this method as well.

\begin{acknowledgements}
This work was supported by French state funds managed by the ANR within the Investissements d'Avenir program 
under references ANR-16-CE33-0009 (DeepSurg), ANR-11-LABX-0004 (Labex CAMI) and ANR-10-IDEX-0002-02 
(IdEx Unistra). 
\end{acknowledgements}

\begin{extranote}
  \textbf{Conflicts of interest} The authors declare that they have no conflict of interest.
  \newline
  \textbf{Ethical approval} This article does not contain any studies with human participants 
  or animals performed by any of the authors.
  \newline
  \textbf{Informed consent} Statement of informed consent was not applicable since the 
  manuscript does not contain any patient data.
\end{extranote}


\bibliographystyle{spmpsci}   
\bibliography{bibliography}

\end{document}